\documentclass[pdflatex,iicol]{sn-jnl}

\jyear{2021}%

\theoremstyle{thmstyleone}%
\theoremstyle{thmstyletwo}%

\theoremstyle{thmstylethree}%

\begin{document}

\title[MA ROM]{Manifold Alignment-Based Multi-Fidelity Reduced-Order Modeling Applied to Structural Analysis}



\author[1]{\fnm{Christian} \sur{Perron}}\email{christian.perron@gatech.edu}
\equalcont{These authors contributed equally to this work.}

\author[1]{\fnm{Darshan} \sur{Sarojini}}\email{sdarshan@gatech.edu}
\equalcont{These authors contributed equally to this work.}

\author*[1]{\fnm{Dushhyanth} \sur{Rajaram}}\email{drajaram6@gatech.edu}
\equalcont{These authors contributed equally to this work.}

\author[1]{\fnm{Jason} \sur{Corman}}\email{jason.corman@gatech.edu}
\equalcont{These authors contributed equally to this work.}

\author[1]{\fnm{Dimitri} \sur{Mavris}}\email{dimitri.mavris@aerospace.gatech.edu}
\equalcont{These authors contributed equally to this work.}

\affil*[1]{\orgdiv{Daniel Guggenheim School of Aerospace Engineering}, 
           \orgname{Georgia Institute of Technology}, \orgaddress{\street{North Avenue NW}, \city{Atlanta}, \postcode{30332}, \state{Georgia}, \country{USA}}}



\abstract{
This work presents the application of a recently developed parametric, non-intrusive, and multi-fidelity reduced-order modeling method on high-dimensional displacement and stress fields arising from the structural analysis of geometries that differ in the size of discretization and structural topology.
The proposed approach leverages manifold alignment to fuse inconsistent field outputs from high- and low-fidelity simulations by individually projecting their solution onto a common latent space.
The effectiveness of the method is demonstrated with two multi-fidelity scenarios involving the structural analysis of a benchmark wing geometry.
Results show that outputs from structural simulations using incompatible grids, or related yet different topologies, are easily combined into a single predictive model, thus eliminating the need for additional pre-processing of the data.
The new multi-fidelity reduced-order model achieves a relatively higher predictive accuracy at a lower computational cost when compared to a single-fidelity model.
}

\keywords{
Multi-fidelity,
Reduced Order Model,
Wing Structural Analysis,
Machine Learning
}


\maketitle

\renewcommand*{\arraystretch}{1.3}

\section{Introduction}\label{sec:intro}

The design and certification of aircraft are increasingly relying on numerical simulation instead of physical experiments owing to the former's relatively rapid turnaround time and high accuracy.
Despite the increased reliance on numerical simulations, routinely encountered many-query scenarios such as design space exploration, optimization, uncertainty quantification, etc., require repetitive evaluation of computationally expensive codes.
Consequently, engineering workflows with expensive numerical simulations \emph{in-the-loop} can suffer from impractical compute times.
Surrogate models have provided respite to this computational burden by trading the high accuracy of physics-based numerical models for faster evaluation times.

While conventional surrogate models have mainly focused on the prediction of scalar quantities~\cite{Queipo2005, Forrester2009, Yondo2018, Bhosekar2018}, reduced-order models (ROMs) have enabled the prediction of spatially and temporally distributed field quantities.
This class of methods reduces the high-dimensional nature of the problem by extracting a low-dimensional subspace (or more generally, latent space) that best captures the most dominant physical features of the solution space.
In a sense, by preserving and capturing the underlying spatial and temporal correlation within the high-dimensional solutions, ROMs can offer physically richer predictions.
This trait renders ROMs useful in enabling rapid multidisciplinary analyses where high-dimensional coupling variables must be passed between simulations to achieve consistency between, for instance, the aerodynamics and the structures disciplines.

The construction of surrogate models for both scalar and field quantities suffers from a major drawback when the underlying response is highly nonlinear and consists of many independent parameters.
In such cases, most methods require a large amount of training data to build a model with reasonable predictive accuracy.
Training data generated from high-fidelity simulations usually incur a substantial computational cost, which may arguably defeat the purpose of constructing the surrogate model.

Compared to scalar quantities, the multi-fidelity combination of fields brings more challenges because of the potential inconsistency between results produced by different analyses~\cite{Perron2020}.
For instance, it is well known that a fine mesh FEM model generally produces more accurate results than a coarse mesh FEM model but at a higher computational cost.
Although fine mesh and coarse mesh FEM models represent the same structure, they have inconsistent field outputs.
This inconsistency non-trivializes the process of combining their high-dimensional results into a single predictive ROM.

Recent work by Perron et al.~\cite{Perron2020, Perron2020a, perron2021multi} introduced a novel ROM method based on manifold alignment (MA-ROM) to overcome the issues caused by inconsistent multi-fidelity field outputs.
The proposed method employs manifold alignment to find a common low-dimensional representation of two datasets composed of heterogeneous high-dimensional field outputs.
The method's capabilities and performance in comparison to other multi-fidelity ROM methods were demonstrated on aerodynamic problems, and the results have shown that multi-fidelity fields having distinct sizes and topologies could be readily combined into a single model with superior predictive accuracy.
In an effort to investigate the effectiveness of MA-ROM against the unique challenges presented by structural simulations, this work primarily focuses on empirically assessing the performance of the MA-ROM by constructing surrogates on high-dimensional outputs from related yet distinct 3D structural analyses.

The Common Research Model (CRM) wing geometry serves as a testbed to showcase the method.
This demonstration aims to construct multi-fidelity ROMs with wing planform shape parameters as input and structural field responses as outputs for two challenging multi-fidelity scenarios.
The first scenario captures the differences in the outputs arising from the coarse and fine mesh FEM representation discussed earlier.
Few samples from the more accurate but expensive fine mesh field are combined with many coarse mesh samples.
The second scenario is a novel demonstration that tackles field outputs from models with inconsistent topologies;~one, a primary dataset from a geometry with a high rib count, and two, an auxiliary dataset from a low rib count wing.
This scenario is designed to test whether the MA-ROM can \emph{warm-start} the training of a prediction model using existing data from a related, yet different problem.
Results show that the proposed MA-ROM can effectively combine field outputs having inconsistent dimensions and topologies from both the scenarios described above.
The method produces relatively more accurate results when applied to the displacement rather than the von-Mises stress field.

The remainder of the manuscript is organized as follows.
Section~\ref{sec:lit} briefly discusses multi-fidelity surrogate models and ROMs in the realm of structural analysis.
Section~\ref{sec:formulation} reviews the conventional POD-based ROM and presents a succinct summary of the multi-fidelity ROM method based on manifold alignment.
Section~\ref{sec:application} introduces the details of the application problem and the two multi-fidelity scenarios considered in this work.
Finally, Section~\ref{sec:results} presents the results of the experiments and discusses the performance of the MA-ROM in comparison to a single-fidelity POD and interpolation-based approach.
\section{Literature Review}\label{sec:lit}

Several authors have adopted a surrogate-based optimization strategy to perform an aero-structural optimization of an aircraft wing.
Abdelkader and Stanislaw~\cite{benaouali2019multidisciplinary} created radial basis function (RBF) models for scalar outputs (e.g., lift, drag, weight, and aggregated structural failure) from the high-fidelity analysis.
Bordogna et al.~\cite{bordogna2017surrogate} created surrogate models trained with lift and drag coefficients from rigid RANS computations as constraints in optimization.
Rasmussen et al.~\cite{rasmussen2009optimization} used a polynomial regression to explore a joined-wing aircraft's optimal regions and design trends.
Joined-wing planform parameters and structural component thicknesses were the inputs to the surrogate model, and the sized total weight was the output.
Cipolla et al.~\cite{cipolla2021doe} created regression models for predicting the wing structural mass, estimating the maximum von-Mises stress in each main wing, and evaluating the highest wing-tip displacement between the front and rear wings.
Surrogate models have also been employed in other many-query applications such as reliability-based optimization of aircraft wing structural components~\cite{li2019reliability,neufeld2010aircraft}, evaluation of certification constraints~\cite{sarojini2020certification}, or uncertainty quantification of dynamic loads on the airframe~\cite{duca2018effects}.

In the literature mentioned so far, surrogate models have been used to approximate scalar quantities as a function of wing planform, geometrical parameters, aerodynamic parameters, and structural thicknesses.
While the wing's weight and global failure criteria provide practical metrics for overall optimization purposes, these aggregated quantities hide many details useful to a designer.
One such detail, for instance, is the location of maximum stress or strain and its spatial shifting as design parameters change. 
Furthermore, in an aero-structural context, the full displacement of a wing structure under loads must be provided to an aerodynamic analysis code in order to adequately compute the aerodynamic forces for a given flight shape.
ROMs also have other applications, such as real-time estimation of a vehicle’s evolving structural state from sensor data~\cite{mainini2015surrogate}.

As for the prediction of structural field quantities, Lieu et al.~\cite{lieu2006reduced,lieu2007adaptation} proposed a POD-based ROM for CFD-based aeroelastic analysis using the F-16 aircraft as a test case.
Mainini and Willcox~\cite{mainini2015surrogate} used the POD to construct ROMs to predict high-fidelity FEM solutions for the estimation of structural capability from sensor data.
ROMs have also been extensively leveraged for the prediction of loads.
Bekemeyer and Timme~\cite{bekemeyer2019flexible} used a subspace projection model reduction to obtain transonic gust loads.
Ripepi et al.~\cite{ripepi2018reduced} created a ROM of the aerodynamic influence coefficients and used steady CFD solutions to obtain loads preserving non-linearities.
ROMs have also been used to reduce the computational cost of performing flutter and limit cycle oscillations analysis~\cite{zhang2012efficient}.
Note that the literature referenced above pertains to single-fidelity ROMs, i.e., constructed using data from one source.

The idea of leveraging a lower-fidelity analysis to reduce the computational cost of generating surrogate models has been shared by multiple authors in the past.
For example, Forrester et al.~\cite{Forrester2007} used multi-fidelity surrogate models for the aerodynamic optimization of a wing, and Toal~\cite{Toal2015} followed a similar process for the optimization of a high-pressure compressor.

Whereas multi-fidelity methods are relatively mature for scalar surrogate models, the prediction of field quantities brings additional challenges, as pointed out in Section~\ref{sec:intro}.
As described by Perron~\cite{Perron2020a}, these discrepancies can be due to differences in grid sizes, grid topologies, and field features.
A few efforts have attempted to avoid this issue by enforcing the same discretization for both the high- and low-fidelity analysis and have achieved computational savings by using a simplified physics model.
This approach was used by Mifsud et al.~\cite{Mifsud2016, Mifsud2008} for the aerodynamic analysis of a projectile and by Bertram et al.~\cite{Bertram2018} for the aerodynamic analysis of a road vehicle.
Alternatively, one could force the high- and low-fidelity field representations to be consistent by mapping all the results on a common grid, assuming that both fidelities have compatible field topologies.
Malouin et al.~\cite{Malouin2013} demonstrated this strategy for the analysis of a transonic airfoil.
Similarly, Benamara et al.~\cite{Benamara2016,Benamara2017} applied this approach to the analysis of a compressor blade.
More recently, Perron et al.~\cite{Perron2020a} proposed the MA-ROM method that leverages manifold alignment~\cite{Ham2005,Wang2009} to resolve discrepancies between the high- and low-fidelity fields.
The effectiveness of this method was tested with the aerodynamic analysis of an airfoil and a wing.
Furthermore, results have shown that even when fields with consistent representation are used (i.e., by mapping results to a common grid), the MA-ROM method offered an accuracy on par or superior to previous multi-fidelity ROM methods~\cite{Perron2020}.

Although multi-fidelity ROM methods have been successfully applied to aerodynamic problems, the viability of these approaches has not yet been established for structural applications.
Moreover, topological differences, in terms of model representation and structural layout, are much more prevalent in structural than in aerodynamic problems.
Previous data suggest that the MA-ROM is especially well suited for dealing with multi-fidelity structural data with possibly inconsistent representations.
Empirically testing this hypothesis is the primary goal of what follows.
\section{Formulation}\label{sec:formulation}

The MA-ROM method can be interpreted as a combination of three techniques: proper orthogonal decomposition (POD), manifold alignment (MA), and hierarchical Kriging (HK).
The following section briefly presents each of these techniques and describes how these are combined to form the MA-ROM method.

\subsection{Proper Orthogonal Decomposition}

The POD method~\cite{Lumley67} is a linear and unsupervised dimensionality reduction technique that is also commonly referred to as principal component analysis (PCA) in the field of machine learning.
It is arguably the most common dimensionality reduction method used in the context of projection-based ROM.

Let us consider the data matrix $\mathbf{X} \in \mathbb{R}^{d\times n}$ that contains $n$ samples $\mathbf{x}_j$ of dimension $d$ organized column-wise.
Let us further assume that $\mathbf{X}$ has been pre-processed such that its sample mean
\begin{equation}
    \overline{\mathbf{x}} = \frac{1}{n}\sum_{j=1}^{n} \mathbf{x}_j
\end{equation}
is zero, meaning that the mean has been subtracted from the original data set beforehand.

The POD method attempts to construct a simpler representation of $\mathbf{X}$ that captures most of the data variance.
This is achieved by finding basis vectors $\boldsymbol{\phi}_j \in \mathbb{R}^d$ (also called POD modes) that maximizes the following \emph{Rayleigh quotient}~\cite{Magnus1997}
\begin{equation}\label{eq:Rayleigh}
    \max_{\boldsymbol{\phi}_j} \frac{\boldsymbol{\phi}_j^T \mathbf{S} \, \boldsymbol{\phi}_j}{\boldsymbol{\phi}_j^T \boldsymbol{\phi}_j}
\end{equation}
where $\mathbf{S}= \frac{1}{n} \mathbf{X}\mathbf{X}^T \in \mathbb{R}^{d \times d}$ is the sample covariance matrix.

The solution of to the above objective is given by the eigendecomposition of $\mathbf{S}$~\cite{DeBie2005} such that
\begin{equation}
    \mathbf{S} = \boldsymbol{\Phi} \boldsymbol{\Lambda} \boldsymbol{\Phi}^T
\end{equation}
where $\boldsymbol{\Phi} \in \mathbb{R}^{d \times d}$ contains the eigenvectors, and $\boldsymbol{\Lambda} \in \mathbb{R}^{n \times n}$ is a diagonal matrix where the entries $\lambda_j$ are the corresponding eigenvalues.
We define the POD basis $\boldsymbol{\Phi}_k \in \mathbb{R}^{d \times k}$ as a low-rank approximation of $\boldsymbol{\Phi}$ whose columns are composed of the $\boldsymbol{\phi}_j$ associated with the $k$-largest $\lambda_j$, where $k \ll d$.

Using $\boldsymbol{\Phi}_k$, the original data can be projected into the POD latent space such that
\begin{equation}
    \mathbf{Z} = \boldsymbol{\Phi}_k^T \mathbf{X}
\end{equation}
where $\mathbf{Z} \in \mathbb{R}^{k \times n}$ contains the latent coordinates $\mathbf{z}_j \in \mathbb{R}^k$ (also known as POD coefficients) corresponding to $\mathbf{x}_j$.
Conversely, the original data can be approximately reconstructed from $\mathbf{Z}$ such that
\begin{equation}
    \Xtilde = \boldsymbol{\Phi}_k \mathbf{Z}
\end{equation}
Note that $\Xtilde \in \mathbb{R}^{d \times n}$ is an approximation of $\mathbf{X}$ since only the $k$-first POD modes are used.

Lastly, the dimension $k$ of the POD basis is usually determined using the relative information content (RIC) criterion~\cite{Pinnau2008,Vendl2013} defined as
\begin{equation}\label{eq:ric}
    \text{RIC} = \frac{\sum_{j=1}^{k} \lambda_j}{\sum_{j=1}^{d} \lambda_j}
\end{equation}
The RIC criterion essentially represents the ratio of the total variance captured by the POD basis.
A higher RIC implies a more accurate representation of $\mathbf{X}$, but typically results in a larger latent space dimension.
Users typically select $k$ such that the RIC is greater than some threshold, and values of 99.9\% or higher are fairly common in the ROM literature.

\subsection{Manifold Alignment}

Computing the POD of different datasets, say a high- and low-fidelity dataset, will clearly result in distinct POD bases and coefficients.
The purpose of manifold alignment~\cite{Ham2005,Wang2009} is to uncover a common latent space, instead of distinct latent spaces obtained via the POD, where disparate data can be more easily compared.
In the current context, the manifold alignment allows us to combine field results having different representations.
Specifically, the MA-ROM method leverages the Procrustes manifold alignment~\cite{Wang2008} method as it can readily be combined with the established POD-based ROM methodology~\cite{Perron2020}.

Consider $\mathbf{X} \in \mathbb{R}^{d \times n}$ and $\mathbf{Y} \in \mathbb{R}^{q \times m}$ as matrices containing the high- and low-fidelity samples $\mathbf{x}_j \in \mathbb{R}^d$ and $\mathbf{y}_j \in \mathbb{R}^q$ respectively.
Note that in view of a realistic scenario, it is assumed there are more low- than high-fidelity samples, i.e., $ m \gg n$.
However, no assumption is made regarding the dimensionality of both sets of fields;~they can be different, i.e., $d \neq q$.
Furthermore, let the first $n$ samples $\mathbf{x}_j$ and $\mathbf{y}_j$ have a pair-wise correspondence;~meaning they are generated by evaluating the respective simulation at identical design parameter values.
Consequently, the low-fidelity data can be partitioned as $\mathbf{Y} = [\mathbf{Y}_\text{L}, \mathbf{Y}_\text{U}]$, where $\mathbf{Y}_\text{L} \in \mathbb{R}^{q \times n}$ contains the solutions \emph{linked} to $\mathbf{X}$, while $\mathbf{Y}_\text{U} \in \mathbb{R}^{q \times (m - n)}$ holds the \emph{unlinked} data.

The first step of the Procrustes manifold alignment is to find the latent space, i.e., the POD basis, of both the high- and low-fidelity datasets.
Let us begin by defining
\begin{align}\label{eq:mf-pod}
    \mathbf{Z} &= \boldsymbol{\Phi}_k^T \mathbf{X}\\
    \mathbf{W} &= \boldsymbol{\Psi}_k^T \mathbf{Y}
\end{align}
where $\mathbf{Z} \in \mathbb{R}^{k \times n}$ and $\mathbf{W} \in \mathbb{R}^{k \times m}$ are the low- and high-fidelity POD coefficients respectively, while $\boldsymbol{\Phi}_k \in \mathbb{R}^{d \times k}$ and $\boldsymbol{\Psi}_k \in \mathbb{R}^{q \times k}$ are the associated POD modes.
Note that despite having the same dimension $k$, $\mathbf{Z}$ and $\mathbf{W}$ still reside in different subspaces.
As with $\mathbf{Y}$, we split the low-fidelity data into $\mathbf{W} = [\mathbf{W}_\text{L}, \mathbf{W}_\text{U}]$, where $\mathbf{W}_\text{L} \in \mathbb{R}^{k \times n}$ and $\mathbf{W}_\text{U} \in \mathbb{R}^{k \times (m - n)}$ are the linked and unlinked low-fidelity POD coefficients.

Then, the POD coefficients of both datasets are aligned using Procrustes analysis~\cite{Gower2010}.
The goal of this step is to find an affine transformation of $\mathbf{W}_\text{L}$ that would optimally align it with $\mathbf{Z}$.
This is expressed by the following objective
\begin{equation}
\begin{split}
    \min_{s, \mathbf{t}, \mathbf{Q}} \quad& \lVert \mathbf{Z} - s \mathbf{Q} (\mathbf{W}_\text{L} -\mathbf{t}) \lVert_F \\
    \text{s.t.} \quad&  \mathbf{Q}^T \mathbf{Q} = \mathbf{I}
\end{split}
\end{equation}
where $s$ is an isotropic scaling factor, $\mathbf{t} \in \mathbb{R}^{k}$ is a translation vector, and $\mathbf{Q} \in \mathbb{R}^{k \times k}$ is an orthogonal transformation matrix.

The optimal $\mathbf{t}$ is obtained by shifting $\textbf{W}_\text{L}$ such that its centroid coincides with that of $\textbf{Z}$.
It is assumed that $\mathbf{X}$ has been centered prior to computing the POD such that the mean of $\mathbf{Z}$ is also zero.
Thus, the optimal $\mathbf{t}$ is given by
\begin{equation}
    \textbf{t} = \frac{1}{n}\sum_{j=1}^{n} \mathbf{w}_j
\end{equation}
For convenience, $\textbf{W}'_\text{L} = \textbf{W}_\text{L} - \textbf{t}$ is defined as the shifted low-fidelity latent variable.

The remaining optimal parameters of the Procrustes analysis are then obtained by the following relations
\begin{align}
    s &= \frac{\text{tr}(\boldsymbol{\Sigma})}{\text{tr}\left(\mathbf{W}'_\text{L} (\mathbf{W}'_\text{L})^T\right)} \\
    \textbf{Q} &= \textbf{V} \textbf{U}^T
\end{align}
In the results shown above, the matrices $\mathbf{U}$, $\boldsymbol{\Sigma}$, and $\mathbf{V}$, are the outputs of the following singular value decomposition (SVD)
\begin{equation}
    \textbf{U} \boldsymbol{\Sigma} \textbf{V}^T = \textbf{W}'_\text{L} \textbf{Z}^T
\end{equation}
Finally, the optimal transformation obtained using $\mathbf{W}_\text{L}$ is applied to all the low-fidelity data $\mathbf{W}$.
The aligned POD coefficients are thus given by
\begin{equation}\label{chap3-eq:z_align}
    \textbf{Z}_\text{lo} = s \, \textbf{Q} \left(\textbf{W} - \textbf{t}\right)
\end{equation}
where $\textbf{Z}_\text{lo} \in \mathbb{R}^{k \times m}$ contains the low-fidelity POD coefficients mapped onto the high-fidelity POD basis.
For the sake of clarity, the high-fidelity data $\textbf{Z}$ are referred to as $\textbf{Z}_\text{hi}$ from here onward.

\subsection{Hierarchical Kriging}

Following the POD and the manifold alignment, the high- and low-fidelity data have been reduced to the latent coordinates $\textbf{Z}_\text{hi}$ and $\textbf{Z}_\text{lo}$.
The multi-fidelity data must then be combined into a regression model in order to predict new field solutions at unseen design parameters.
In practice, any multi-fidelity regression technique could be used for this purpose and the literature contains multiple options such as bridge functions~\cite{Choi2009, Eldred2004, Tang2005} and cokriging~\cite{Myers1982,Kennedy2000,Han2012}.
In this study, the hierarchical Kriging (HK) method~\cite{Han2012a} is used to remain consistent with previous MA-ROM studies~\cite{Perron2020,Perron2020a,Decker2021}.

As defined by Han and G\"{o}rtz~\cite{Han2012a}, the HK formulation is nearly identical to cokriging~\cite{Kennedy2000}, although the former is marginally simpler to implement and provides a smoother posterior variance prediction.
Given a high-fidelity function $g_\text{hi}: \mathbf{p} \mapsto z_\text{hi}$, where $\mathbf{p} \in \mathbb{R}^b$ is a design parameter vector and $z_\text{hi}$ is a high-fidelity response, the HK predictor takes the following form
\begin{equation}\label{eq:hk-predictor}
    \gtilde_\text{hi}(\mathbf{p}) = \beta\,\gtilde_\text{lo}(\mathbf{p}) + \mathbf{w}^T \mathbf{r}(\mathbf{p})
\end{equation}
where $\beta$ is a multi-fidelity scaling factor, $\mathbf{w} \in \mathbb{R}^n$ is a weight vector, and $\mathbf{r}(\mathbf{p})$ is a kernel function.
In this example, low-fidelity information is included in the form of $\gtilde_\text{lo}(\mathbf{p})\mathbf{p} \mapsto \tilde{z}_\text{lo}$, which is a Kriging predictor of some low-fidelity response.
Loosely speaking, the HK predictor of equation~\eqref{eq:hk-predictor} can be viewed as a Kriging model where the response trend is represented by $\beta\,\gtilde_\text{lo}(\mathbf{p})$.
Note that for the current study, both $\gtilde_\text{hi}(\mathbf{p})$ and $\gtilde_\text{lo}(\mathbf{p})$ use a Mat\'ern~3/2 kernel and the model parameters are obtained with a maximum likelihood approach.

In the context of MA-ROM, the HK model is used to predict a new latent coordinate vector $\textbf{z}_\text{hi}$ given some design parameters $\mathbf{p}$.
Since each coordinate contains $k$ individual components, an equal number of regression models $\gtilde_\text{hi}(\mathbf{p})$ must be trained in order to predict the entire $\textbf{z}_\text{hi}$ vector.
The prediction of the high-dimensional field output is then given by
\begin{equation}
    \xtilde = \sum_{i=1}^{k} \boldsymbol{\phi}_i \, \gtilde_{i,\text{hi}}(\mathbf{p})
\end{equation}
where $\gtilde_{i,\text{hi}}(\mathbf{p})$ is the HK model predicting the $i$-th coordinated in the POD latent space associated with the POD mode $\boldsymbol{\phi}_i$.

\subsection{MA-ROM Formulation}

Consider a high- and low-fidelity models defined as $f_x: \mathbf{p} \mapsto \mathbf{x}$ and $f_y: \mathbf{p} \mapsto \mathbf{y}$ respectively.
The MA-ROM formulation can be divided into an \emph{offline} phase where the model is trained, and an \emph{online} phase where predictions are made using the generated model.
The offline phase is summarized through the following steps
\setlist[enumerate,1]{label={\arabic*.}}
\begin{enumerate}
\item \textbf{Generate Linked Data:} Select a set of $n$ distinct designs $\mathbf{P}_\text{L} = [\mathbf{p}_1, \dots, \mathbf{p}_n] \in \mathbb{R}^{b\times n}$ for the linked data.
    Then, sample both $f_x$ and $f_y$ to generate the matrices $\mathbf{X}$ and $\mathbf{Y}_\text{L}$.

    \item \textbf{Generate Unlinked Data:} Select a set of $m - n$ additional designs $\mathbf{P}_\text{U} = [\mathbf{p}_{n+1}, \dots, \mathbf{p}_m] \in \mathbb{R}^{b\times (m-n)}$ for the unlinked data.
    Ideally, $\mathbf{P}_\text{U}$ should not repeat points in $\mathbf{P}_\text{L}$.
    Then, sample only $f_y$ with $\mathbf{P}_\text{U}$ to generate the matrix $\mathbf{Y}_\text{U}$.

    \item \textbf{Align Manifolds:} Apply the Procrustes manifold alignment on $\mathbf{X}$, $\mathbf{Y}_\text{L}$, and $\mathbf{Y}_\text{U}$, to obtain $\mathbf{Z}_\text{hi}$ and $\mathbf{Z}_\text{lo}$. This is divided into the following three sub-steps:
    \begin{enumerate}
        \item Use POD to extract $\boldsymbol{\Phi}_k$ and $\boldsymbol{\Psi}_k$, as well as $\mathbf{Z}_\text{hi}$ and $\mathbf{W}$.

        \item Perform Procrustes analysis on $\mathbf{Z}_\text{hi}$ and $\mathbf{W}_\text{L}$ to evaluate the optimal value of $s$, $\mathbf{t}$, and $\mathbf{Q}$.

        \item Apply the optimal transformation to $\mathbf{W}$ to compute $\mathbf{Z}_\text{lo}$.
    \end{enumerate}

    \item \textbf{Fit HK Model:} Train a set of $k$ HK models $\gbtilde(\mathbf{p}) = [\gtilde_1(\mathbf{p}), \dots, \gtilde_k (\mathbf{p})]$.
    These models combine the information from both $\mathbf{Z}_\text{hi}$ and $\mathbf{Z}_\text{lo}$, together with their associated design parameters $\mathbf{P} = [\mathbf{P}_\text{L}, \mathbf{P}_\text{U}] \in\mathbb{R}^{b\times m}$.
\end{enumerate}
It should be noted that in this study, the design samples in steps 1 and 2 are obtained using Latin hypercube sampling (LHS)~\cite{Santner2003}.
As for the online phase, the process is identical to most non-intrusive ROMs and is given by
\begin{enumerate}[resume*]
    \item \textbf{Predict Latent Variable:} Given an out-of-sample design parameter $\mathbf{p}^*$, predict the corresponding latent variable $\ztilde^*$ using $\gbtilde(\mathbf{p})$.

    \item \textbf{Reconstruct Field:} Using $\boldsymbol{\Phi}_k$ and $\ztilde^*$, compute the predicted value of the new high-fidelity field $\xtilde^*$.    
\end{enumerate}

\section{Application Problem}\label{sec:application}

The application problem considers the analysis of a wing structure to demonstrate the relevance of MA-ROM for large-scale problems representative of an industrial design process. 
The baseline geometry considered for this application problem is the wing of the NASA CRM~\cite{kenway2014aerostructural}.
The CRM is representative of a transonic wide-body transport aircraft. 

\subsection{Aero-Structural Analysis}\label{subsec:crm_baseline}

\begin{figure*}[]
    \centering
    \includegraphics[width=1.8\columnwidth]{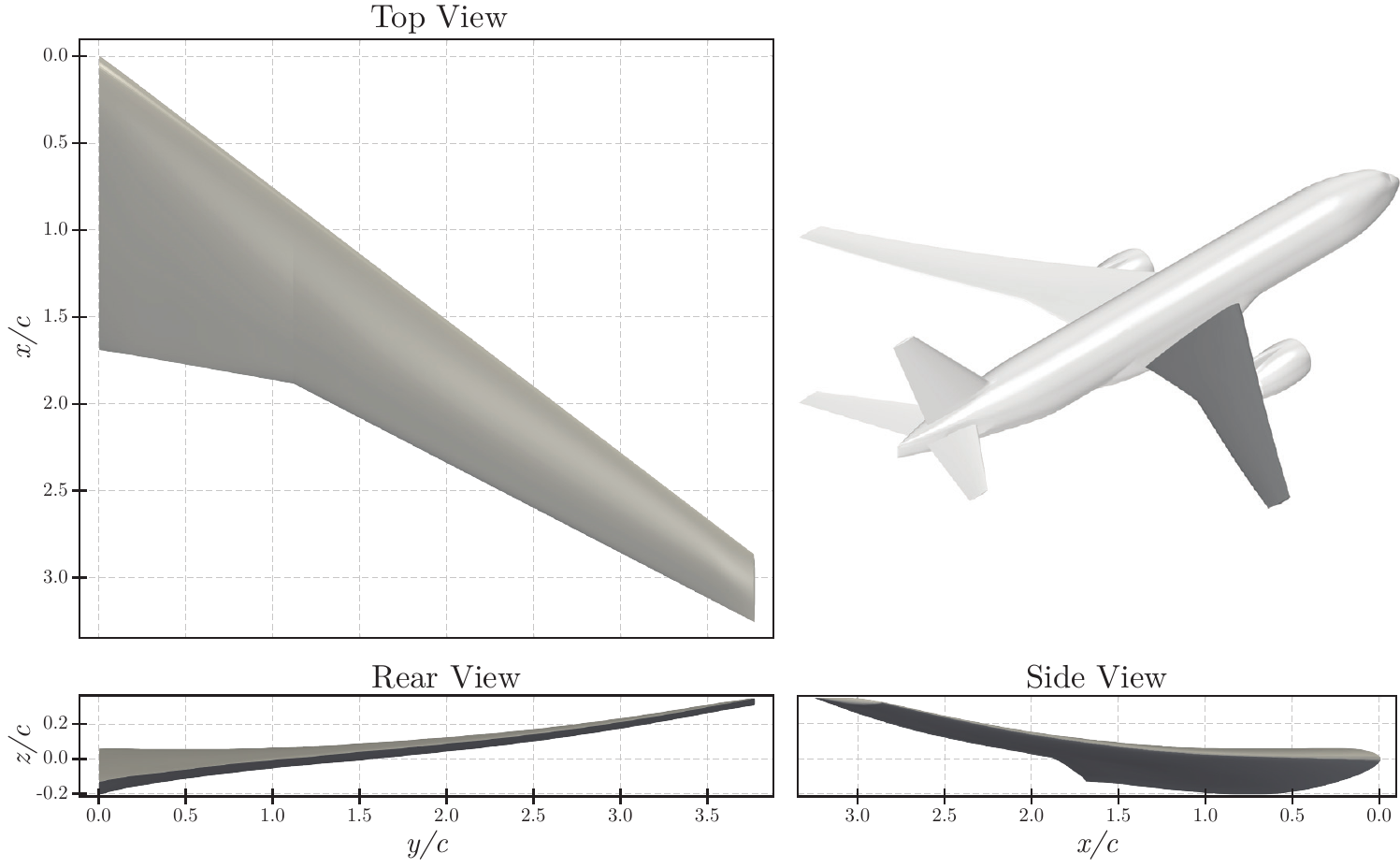}
    \caption{Top, rear, and side view of the CRM wing. The dimensions provided are normalized by the mean aerodynamic chord whose value is 275.8~in. The top right isometric view shows the wing in context of the full aircraft configuration.}
    \label{fig:crm_wing_oml}
\end{figure*}

The baseline OML\footnote{An OpenVSP file of the baseline CRM OML is included with this paper} and flight conditions considered are given in Table~\ref{tab:crm_baseline_oml_fc} and shown in Figure~\ref{fig:crm_wing_oml}. 

\begin{table}[ht]
        \centering
        \caption{Baseline OML \& flight condition values}\label{tab:crm_baseline_oml_fc}
        \begin{tabular}{lcc}
            \hline
            \textbf{Parameter}  & \textbf{Value} & \textbf{Unit}    \\
            \hline\hline
            Planform area       & 4444.38127     & \si{\square\ft}  \\
            Span                & 193.34768      & \si{\ft}         \\
            Aspect ratio        & 8.3325         & -                \\
            Taper ratio         & 0.86031        & -                \\
            Leading edge sweep  & 37             & \si{\deg}        \\
            TOGW                & 650000         & \si{\lb}         \\
            Altitude            & 30000          & \si{\ft}         \\
            Mach number         & 0.8            & -                \\
            \hline
        \end{tabular}
\end{table}

\begin{figure}[h!]
    \centering
    \includegraphics[width=\columnwidth]{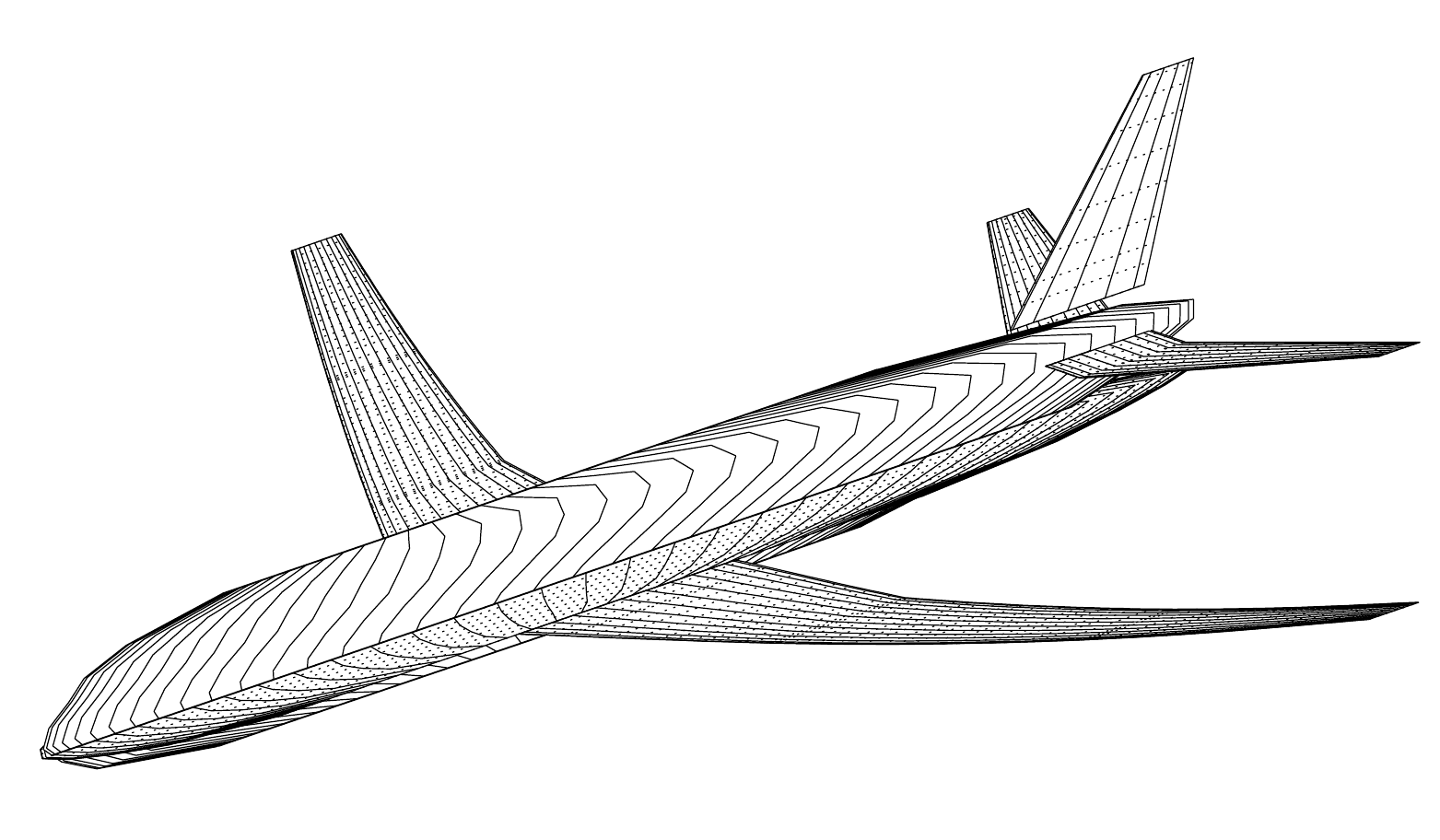}
    \caption{Panel geometry and discretization for CRM AVL model}
    \label{fig:pegasus_avl}
\end{figure}

The aerodynamic loads on the wing are obtained from a panel representation of the aircraft OML.
This panel model has the associated sectional airfoils, local incidence angles, etc. that can be used for creating lower order aerodynamic models for tools like AVL.
The baseline panel geometry for the AVL model
\footnote{The AVL model of the baseline CRM OML is provided as supplementary material.} is shown in Figure~\ref{fig:pegasus_avl}.

\begin{figure}[htb]
    \centering
    \includegraphics[width=\columnwidth]{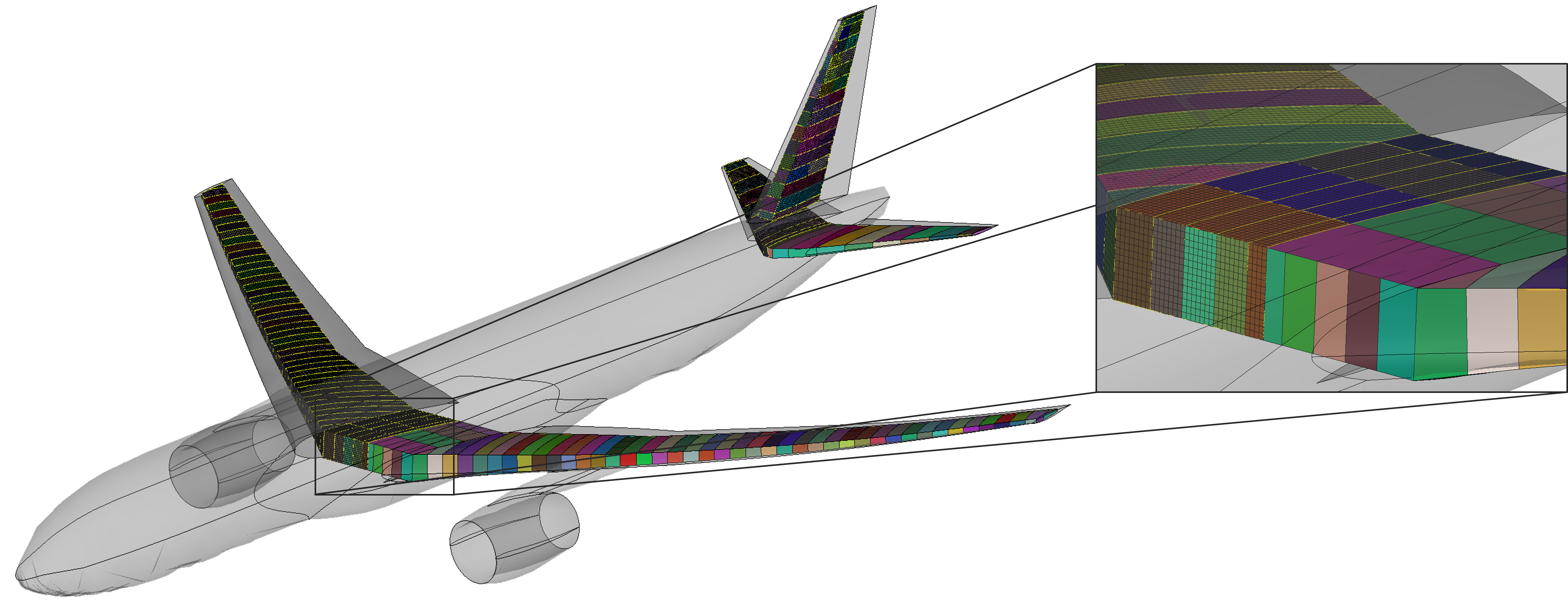}
    \caption{Geometry and mesh of the wing and empennage of the CRM}
    \label{fig:crm_mesh}
\end{figure}

Figure~\ref{fig:crm_mesh} shows the structural geometry layout and mesh for the wing, horizontal tail, and vertical tail within a fused representation of the OML. In this work, we consider only the wing whose structural topology values are given in Table~\ref{tab:crm_baseline_strcTop}.

\begin{table}
        \centering
        \caption{Baseline structural topology values (cn: chord normalized and sn: span normalized).}\label{tab:crm_baseline_strcTop}
        \begin{tabular}{lcc}
            \hline
            \textbf{Parameter}      &   \textbf{Value} & \textbf{Unit} \\
            \hline\hline
            Front spar location     &   0.17           & cn            \\
    		Rear spar location      &   0.68           & cn            \\
    		Rib spacing             &   32             & \si{\in}      \\
    		Rib roll                &   0              & \si{\deg}     \\
    		Rib yaw                 &   -10            & \si{\deg}     \\
    		Root rib location       &   0.1            & sn            \\
    		Tip rib location        &   0.995          & sn            \\
            \hline
        \end{tabular}
\end{table}

The structure is modeled using CQUAD4 shell elements in Nastran\footnote{The Nastran model (i.e., the BDF file) of the baseline CRM OML is provided as supplementary material.} for the ribs, spars, and skins. 
All shell panels are assigned aluminium material with a density of \num[round-mode=places,round-precision=3]{2698.79} \si{\kg\per\meter^3}, a Young's Modulus of \num[round-mode=places,round-precision=3]{71016000118} \si{\newton\per\meter^2}, and a Poisson's ratio of $0.33$.
The following assumptions are made in the structural model:
\begin{enumerate}
    \item No stiffeners and stringers
    \item The thicknesses of the panels are fixed. The values are obtained by averaging thicknesses along the span for a baseline sizing case where the CRM was sized for 2.5g and -1.0g static maneuvers at flight conditions given in Table~\ref{tab:crm_baseline_oml_fc}.
    \begin{itemize}
        \item Front spar: \num[round-mode=places,round-precision=3]{0.465988771929824} \si{\in}
        \item Rear spar: \num[round-mode=places,round-precision=3]{0.257907054747054} \si{\in}
        \item Upper skin: \num[round-mode=places,round-precision=3]{0.923677840755735} \si{\in}
        \item Lower skin: \num[round-mode=places,round-precision=3]{0.657111039136302} \si{\in}
        \item Rib: \num[round-mode=places,round-precision=3]{0.148520056980057} \si{\in}
    \end{itemize}
\end{enumerate}
These assumptions are merely simplifications to make the modeling easier.
The proposed multi-fidelity method's performance is neither affected nor depends on this simplification.
Stiffeners can be included by using the smeared-stiffener calculations as done by Kennedy and Martins~\cite{kennedy2014parallel}. 

The transfer of aerodynamic loads from rigid-wing aerodynamics in AVL to the structure is done by integrating pressure differentials for elements on the aerodynamic panels to reference points along the span of a wing-like component.
The centers of mass of the ribs are chosen as reference points along the span of the wing. 
The forces and moments associated with each reference point are then distributed to the structure via rigid body elements (RBEs) at key intersecting nodes between the ribs and skin.

The structure is analyzed in Nastran using SOL101. Field outputs such as internal loads, stresses, strains, and displacements are obtained.
The proposed MA-ROM approach is used to obtain a surrogate of these field outputs as a function of the input design variables that will be described in Section~\ref{subsec:dv_parametrization}.

It should be noted that in this work each shell element has an assumed thickness.
The Nastran model can be used to size the structure for the applied loads using Hypersizer~\cite{solano2020structural,solano2021pegasus}.
Other fields can be generated from Hypersizer outputs such as optimal sectional dimensions (skin thickness, stiffener height, etc.) and margins of safety for multiple failure criteria.

\subsection{Parametrization}\label{subsec:dv_parametrization}

In the proposed problem, global wing parameters are considered for the design parametrization.
These include the wing sweep, dihedral, tip twist, and span as shown in Table~\ref{tab:ffd-param}.
Note that the bounds are represented as percentage increments/decrements with respect to the original aircraft OML.

\begin{table}
        \centering
        \caption{Design parameters and bounds considered for the application problem. Values are increments to the original CRM geometry}\label{tab:ffd-param}
        \begin{tabular}{lc}
            \hline
            \textbf{Parameter}      &   \textbf{Bounds} \\
            \hline\hline
            Wing sweep     &   $\pm 10^\circ$ \\
    		Wing dihedral      &   $\pm 10^\circ$  \\
    		Wing tip twist &   $\pm 10^\circ$    \\
    		Wing span &   $\pm 20\%$     \\
            \hline
        \end{tabular}
\end{table}

While the MA-ROM approach can combine fields of high- and low-fidelity having different representations, it still requires each fidelity level to internally use a consistent representation.
For instance, a similar grid topology, in terms of size and connectivity, should be used for all the high-fidelity results, and likewise for the low-fidelity results.
However, this grid topology does not need to be between both high- and low-fidelity results since any discrepancy between fidelity levels will be addressed with the manifold alignment.
For the aforementioned reason, a grid deformation approach is used to apply the design parameters of Table~\ref{tab:ffd-param} on both the structural and aerodynamic grids.

\begin{figure}[htb]
    \centering
    \includegraphics[width=0.8\columnwidth]{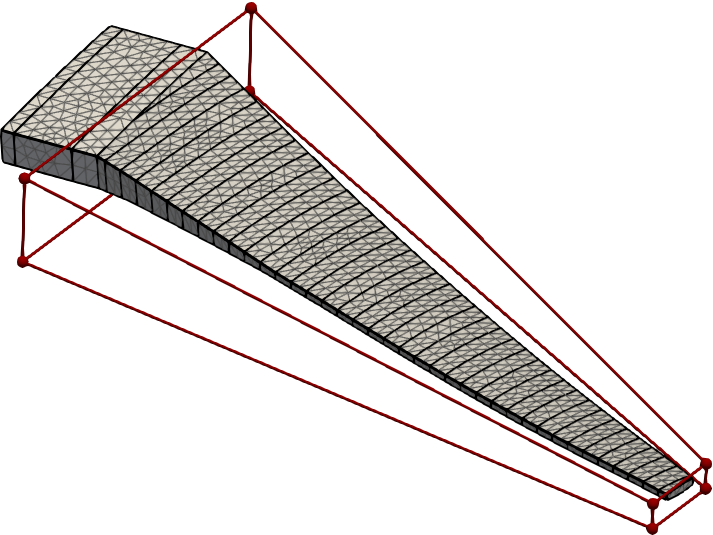}
    \caption{Structural grid of the CRM wing encased in an FFD box (red).}
    \label{fig:ffd-crm}
\end{figure}

In the current study, the grid is deformed using a free-form deformation (FFD) approach~\cite{Kenway2010}.
This technique encases a geometry inside a lattice or box as shown in Figure~\ref{fig:ffd-crm}.
The FFD box behaves as a flexible rubber-like material such that displacement of the boundary nodes is propagated to the geometry within.
For instance, given the current problem, the wing sweep is modified by translating backward or forward all the end nodes of the FFD box.
In addition to maintaining the grid topology, the FFD techniques can effectively be applied to any discretized geometry and without the need for CAD software.
Note that this technique is popular in the context of aerodynamic shape optimization, some examples of which can be found in Lyu et al.~\cite{Lyu2014} and Koo and Zing~\cite{Koo2016}. 

\subsection{Multi-Fidelity Scenarios}

The usefulness of the MA-ROM method is demonstrated using two multi-fidelity scenarios based on the CRM test case with an FFD parametrization described earlier.
Each of these scenarios is aimed at combining multi-fidelity fields having different types of inconsistencies.

\subsubsection{Scenario~1: Inconsistent Grid Sizes}\label{sec:scn-1}

\begin{figure}[htb]
    \centering
    \includegraphics[width=0.7\columnwidth]{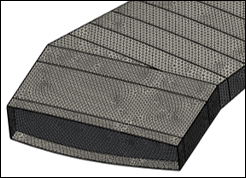}
    
    A) High-fidelity grid (79,167 elements)\vspace{0.1in}
    
    \includegraphics[width=0.7\columnwidth]{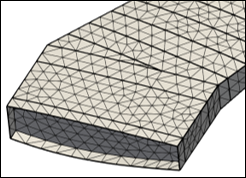}
    
    B) Low-fidelity grid (5,062 elements)
    
    \caption{Comparison the (A) high- and (B) low-fidelity grids used for Scenario~1.}
    \label{fig:scn-1:grid}
\end{figure}

In this scenario, structural results computed on a fine grid are enhanced with data obtained from a similar analysis, yet using a coarser grid.
Specifically, the high- and low-fidelity simulation use grids with a resolution of roughly 79,000 and 5,000 elements respectively as shown in Figure~\ref{fig:scn-1:grid}.
Using a coarser grid, the computational cost of the low-fidelity analysis is reduced from 5.5 to 0.6 CPU-sec on a modern desktop, which represents an order of magnitude reduction.
In this scenario, the purpose of the multi-fidelity method is to reduce the overall training cost of the model by leveraging a computationally cheaper tool.
As such, the multi-fidelity method's predictive performance will be contrasted with that of a single-fidelity model constructed using the results from the fine grid exclusively.
This comparison will look at these models' accuracy for similar training costs, which for the multi-fidelity model, includes the cost of generating both the high- and low-fidelity data.

\subsubsection{Scenario 2: Inconsistent Topologies}\label{sec:scn-2}

\begin{figure}[htb]
    \centering
    \includegraphics[width=0.7\columnwidth]{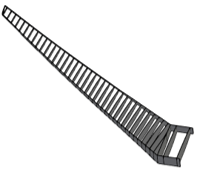}
    
    A) Main structure (high-rib count)\vspace{0.1in}
    
    \includegraphics[width=0.7\columnwidth]{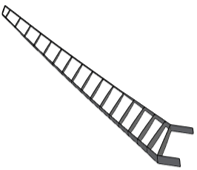}
    
    B) Auxiliary structure (low-rib count)
    
    \caption{Comparison the structural layouts used for the (A) main and (B) auxiliary datasets in Scenario 2.}
    \label{fig:scn-2:layout}
\end{figure}

For the second scenario, the MA-ROM method is used to combine results from similar structural simulations, yet using different structural layouts as shown in Figure~\ref{fig:scn-2:layout}.
Unlike Scenario~1, both simulations are solved using grids with a similar discretization, and as a result, do not follow the usual high- and low-fidelity breakdown.
However, for this scenario, we assume that the results for one of the two structural layouts are readily available in abundance.
For instance, these can correspond to the results previously generated for a different aircraft, or similarly, generated for an earlier version of the aircraft at a point where a different layout was considered.
Therefore, the purpose of the multi-fidelity approach is to \emph{jump-} or \emph{warm-start} the surrogate model generation of a new structural layout using existing data from a different, yet similar application.
Instead of a high- and low-fidelity dataset, the training data is separated into a main and an auxiliary dataset, the former representing the fields that must be predicted by the MA-ROM.
In this example, the main dataset corresponds to the same structural layout used in Scenario~1, and the auxiliary dataset is a structure for the CRM wing using a lower rib count.
The intent behind this scenario is to demonstrate the capability of the MA-ROM to combine structural results despite having inconsistent field topologies.

\subsection{Performance Metrics}\label{sec:metrics}

To establish the benefits of the MA-ROM method for structural problems, its predictive performance is compared to that of a conventional single-fidelity ROM.
This performance is measured in terms of prediction error and is contrasted with the computational cost required to train the model.
As with any empirical model, one can expect MA-ROM to be more accurate when provided with more data, but for it to be cost-effective, it should achieve this accuracy with lesser resource consumption than a single-fidelity method.

\subsubsection{Field Prediction Error}

The prediction error of a ROM or MA-ROM model is typically measured as the difference between the predicted and the actual field responses on previously unseen data, i.e., on data that was left out of the training process.
Using $n_t$ test samples held out from the training data, we define the field prediction error of some field $\mathbf{x} \in \mathbb{R}^d$ as
\begin{equation}
    E(\mathbf{x}) = \sqrt{\frac{\sum_{j=1}^{n_t} \lVert \mathbf{x}^*_j - \widetilde{\mathbf{x}}_j \rVert_2^2}{n_t}}
\end{equation}
where $\widetilde{\mathbf{x}}_j \in \mathbb{R}^d$ is the model prediction from either a ROM or MA-ROM model, and $\mathbf{x}^*_j \in \mathbb{R}^d$ is the corresponding exact solution.
We use the $L^2$-norm of the error field to provide an integrated measure of the prediction error for a given sample.
The test set sample errors are then aggregated using a root-mean-squared operation to represent the overall prediction error of a given model.

Note that the magnitude of $E(\textbf{x})$ is dependent on the field quantity being considered.
To compare errors between different design problems, it is useful to define a normalized prediction error such as
\begin{equation}\label{eq:norm-pred-error}
    \widehat{E}(\textbf{x}) = \frac{E(\textbf{x})}{\sqrt{\sum_{j = 1}^{n_t} \lVert \mathbf{x}_j^* - \overline{\mathbf{x}}\rVert^2_2/n_t}}
\end{equation}
It is worth noting that the normalization term used in equation~\eqref{eq:norm-pred-error} is essentially the total standard deviation of the testing dataset.
As a result, the definition of $\widehat{E}(\textbf{x})$ represents the ratio between the variation of $\textbf{x}$ unexplained by the ROM and the total variation of the data.

\subsubsection{Computational Training Cost}

The computational cost of a ROM can be separated into its evaluation and training costs, which are also referred to as the online and offline costs.
The former is designated as an indicator of the cost of predicting new results and is typically negligible, and as such, we only focus on and report the training cost.
In comparison, the time needed to construct the ROM or MA-ROM model once the data is available is relatively large.

The offline computational cost can be measured in terms of CPU time, usually in CPU-hr or CPU-sec, which is the total amount of computational time used by each processing unit involved.
The CPU time can then be divided by the number of CPUs available to obtain an estimate of the wall-time, neglecting any performance loss from the parallelization.
The cost of each sample is assumed to be a constant in this work, and for Scenario~1 for instance, it is estimated at 5.4402 and 0.5998 CPU-sec for the baseline and coarse grid respectively\footnote{
CPU times are based on an Intel Core i9-10885H CPU.
}.
The total training cost of ROMs in this study is then taken as the number of training samples times the constant cost per sample.

Note that the computational cost can also have a monetary meaning as most high-performance computing facilities provide a price per CPU-hr used.
This price is a result of the operation, maintenance, and acquisition costs of a given supercomputer.
For instance, at the time of writing this manuscript, the Phoenix cluster at the Georgia Institute of Technology offered a preferential rate of \$0.0068/CPU-hr for internal users~\cite{PACE}.

\section{Results}\label{sec:results}

This section presents the predictive performance of the MA-ROM method applied to the CRM wing application problem.
The results for both Scenarios~1 and 2, i.e., combining inconsistent grid sizes and structural topologies respectively, are presented and discussed separately.
For each of the fidelity levels considered, a data set of 1,000 cases
\footnote{The field data generated with each fidelity level are provided as supplementary material.}
is generated using an optimized Latin hypercube sampling method~\cite{Santner2003} and the design bounds defined in Table~\ref{tab:ffd-param}.
From the high-fidelity dataset, a subset of $n$ cases are selected to form the high-fidelity training data.
Afterward, the same $n$ cases and an additional $m - n$ cases are randomly selected from the auxiliary dataset to form the low-fidelity training set.
Once the MA-ROM model is trained, the remainder of the high-fidelity data is used to compute the model prediction error.
The accuracy of the MA-ROM prediction is also compared to that of a conventional single-fidelity ROM, where the latter is only trained with high-fidelity data.

It should be noted that for a given $n$, the above training process is repeated at least 100 times with different subsets of the overall data, and the prediction errors of all models are averaged and reported.
The purpose of these repetitions is to decouple the measured error from the choice of training samples.
In doing so, presented results represent the expected accuracy of the MA-ROM (or single-fidelity ROM) method rather than the performance of one specific model instance.

In the current study, the field outputs of interest are the von-Mises stress distribution $\sigma_\text{vm}$ and the structural displacement $\delta$, although the MA-ROM could be applied to any field responses.
Furthermore, high-fidelity training sample sizes $n$ between 10 and 800 are considered, leaving at least 200 cases for the error calculation.
As for the low-fidelity training set, sample size $m$ is kept proportional to $n$ using the multi-fidelity ratio $\tau = m/n$, and ratio of $\tau = \{2, 4, 8\}$ are considered.
Lastly, all POD bases are computed with an RIC of 99.9999\% or greater for each fidelity level.

\subsection{Scenario~1: Inconsistent Grid Sizes}\label{sec:results:scn-1}

\begin{figure*}[htb]
    \centering
    \includegraphics{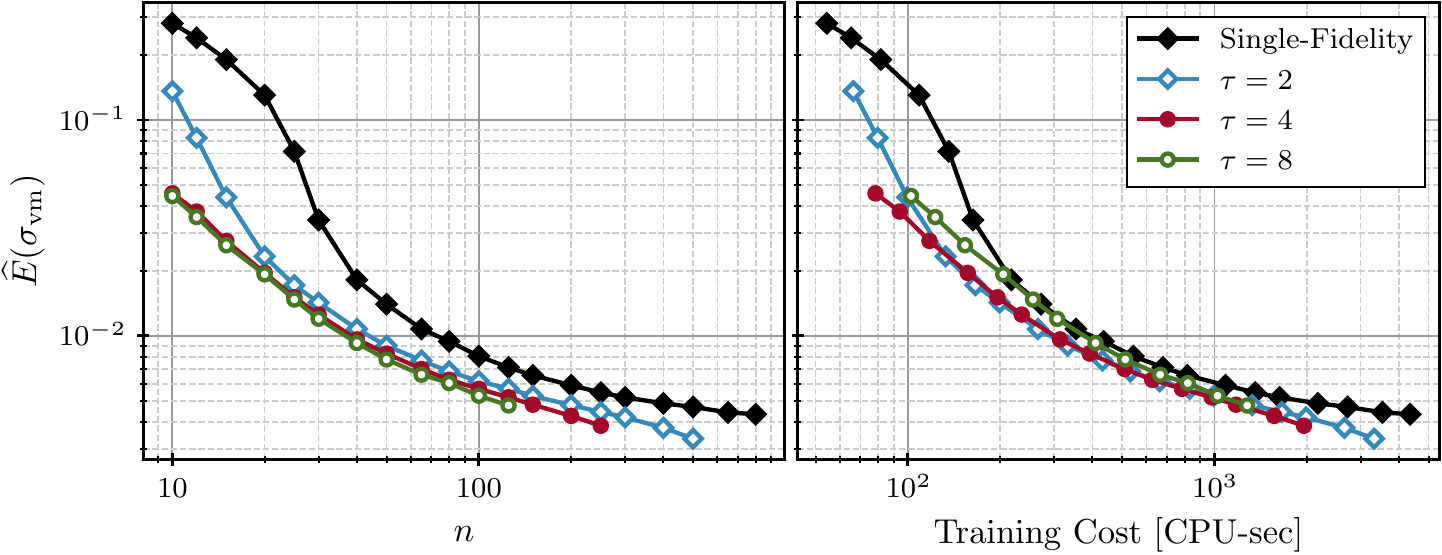}
    \caption{Prediction errors of the von-Mises stress $\sigma_\text{vm}$ using the MA-ROM method to combine fine and coarse grid results.}
    \label{fig:scn-1:stress}
\end{figure*}

\begin{figure*}[htb]
    \centering
    \includegraphics{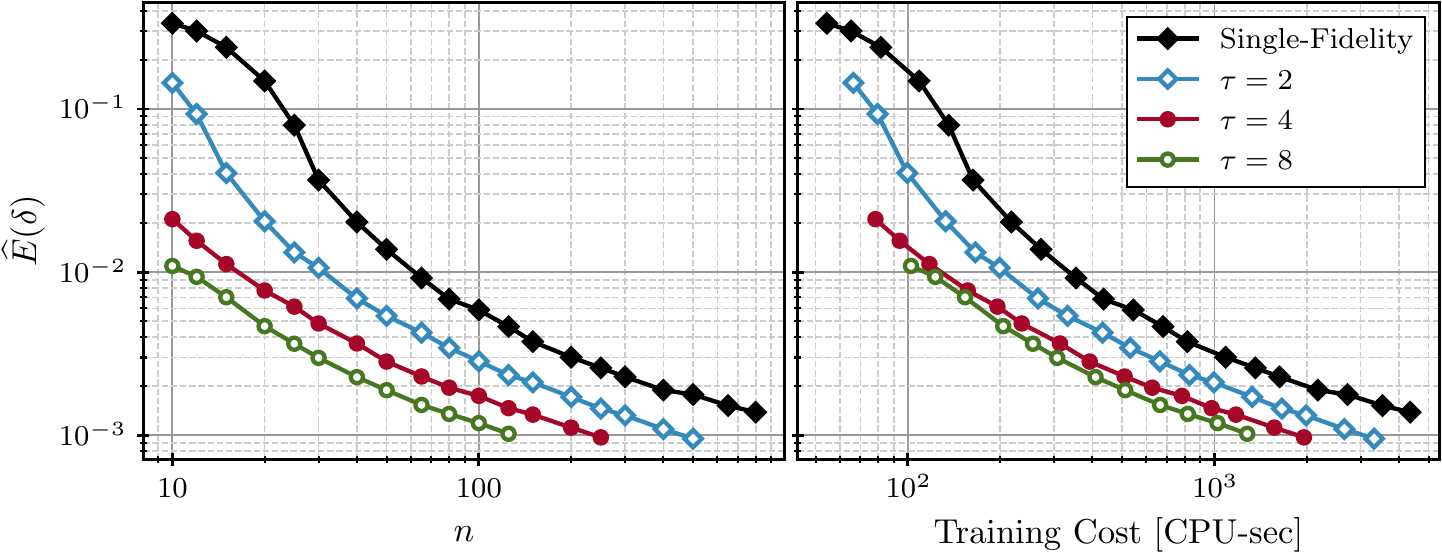}
    \caption{Prediction errors of the structural displacement $\delta$ using the MA-ROM method to combine fine and coarse grid results.}
    \label{fig:scn-1:displacement}
\end{figure*}

As defined in Section~\ref{sec:scn-1}, the first multi-fidelity scenario combines the structural results of the CRM wing obtained from a fine and coarse grid.
The normalized prediction errors on the von-Mises stress distribution $\sigma_\text{vm}$ are presented in Figure~\ref{fig:scn-1:stress} for this scenario.
The blue, red, and green lines represent the results using a multi-fidelity ratio $\tau$ of 2, 4, and 8, respectively.
For comparison, the results for a single-fidelity ROM trained with only the fine grid results are shown using a black line.
The left side of Figure~\ref{fig:scn-1:stress} presents the errors as a function of $n$, while the right side shows the results scaled with the overall training cost of the various models in CPU-sec.
Note that for the multi-fidelity models, the training cost is comprised of the cost for both the high- and low-fidelity training data.
Following a similar presentation, Figure~\ref{fig:scn-1:displacement} also presents the prediction errors of the MA-ROM and ROM methods, although for the structural displacement $\delta$ instead of $\sigma_\text{vm}$.
For the reader's convenience, some of the data points in Figures~\ref{fig:scn-1:stress} and~\ref{fig:scn-1:displacement} are also tabulated in Table~\ref{tab:scn-1}.

\begin{table}[]
    \centering
    \caption{Prediction error of the MA-ROM method applied to Scenario~1.}
    \label{tab:scn-1}
    \begin{tabular}{ccccc}
    \hline
    $n$ & $m$ & $\widehat{E}(\sigma_\text{vm})$ & $\widehat{E}(\delta)$ & CPU-sec \\
    \hline\hline
    100$^*$ & -     & 0.805\%   & 0.587\%   & 544.0 \\
    100     & 200   & 0.616\%   & 0.284\%   & 664.0 \\
    100     & 400   & 0.568\%   & 0.174\%   & 783.9 \\
    100     & 800   & 0.528\%   & 0.119\%   & 1,023.9 \\
    \hline
    183$^*$ & -     & 0.609\%   & 0.321\%   & 1,000.0 \\
    150     & 300   & 0.526\%   & 0.210\%   & 1,000.0 \\
    127     & 508   & 0.514\%   & 0.145\%   & 1,000.0 \\
    97      & 779   & 0.536\%   & 0.120\%   & 1,000.0 \\
    \hline
    \multicolumn{5}{l}{$^*$\footnotesize{Results for a single-fidelity ROM}}
    \end{tabular}
\end{table}

As can be expected, both Figures~\ref{fig:scn-1:stress} and~\ref{fig:scn-1:displacement} show that the accuracy of both the single- and multi-fidelity methods monotonically improves with additional training data.
More importantly, these figures also demonstrate that for all $n$ values, the MA-ROM outperforms the single-fidelity ROM by enhancing the high-fidelity data with additional low-fidelity results.
When compared to a single-fidelity model trained with the same computational budget, say $1,000$~CPU-sec, a relative reduction in the error of up to 15.6\% and 62.6\% is observed for the prediction of $\sigma_\text{vm}$ and $\delta$ respectively as shown in Table~\ref{tab:scn-1}.
However, this improvement appears to depend on the specific field quantity of interest the MA-ROM is applied to.

For the von-Mises stress field, Figure~\ref{fig:scn-1:stress} shows a reasonable reduction in $\widehat{E}(\sigma_\text{vm})$ with $\tau=2$, yet increasing $\tau$ further only provides a marginal improvement.
Specifically, for $n=100$, the normalized prediction error is 0.630\% and 0.574\% for $m=200$ and $800$ respectively, which represents a relative improvement of only 8.9\% for almost double the cost.
When including the additional computational cost of the low-fidelity data into the analysis, the right side of Figure~\ref{fig:scn-1:stress} indicates that for a given prediction error, the MA-ROM can reduce the overall training cost of the model over a single-fidelity approach.
Although, the results also show that predicting $\sigma_\text{vm}$, using $\tau > 4$ is likely not cost-effective.
This suggests that the high- and low-fidelity fields don't share a strong statistical dependence and that the transfer of low-fidelity information within the multi-fidelity method saturates.

On the other hand, for the prediction of the structural displacement, the results of Figure~\ref{fig:scn-1:displacement} exhibit a more pronounced decrease in $\widehat{E}(\delta)$ with additional low-fidelity data.
For $n=100$, the normalized prediction error is 0.275\% and 0.112\% for $m=200$ and $800$ respectively, which represents a relative improvement of 59.3\%.
Even with the additional cost of the low-fidelity samples, the right side of Figure~\ref{fig:scn-1:displacement} presents a clear reduction in $\widehat{E}(\delta)$ for a given computational cost, or equivalently, a reduction in cost for a target prediction error.
While the results show reduced benefits with larger $\tau$, using more low-fidelity data still appears to be cost-effective, especially when compared to the single-fidelity results.

The better performance of the MA-ROM method for the prediction of the displacement field than for the von-Mises stress field can be explained by the higher non-linearity and complexity of the latter.
As one can expect, this suggests that the effectiveness of the multi-fidelity method ultimately depends on the application.
That being said, the results demonstrate that MA-ROM can be a suitable method to predict structural results and has the potential to offer a lower training cost than a single-fidelity method.
Furthermore, these benefits are achieved despite the high- and low-fidelity fields being represented on grids with different discretizations.

\subsection{Scenario~2: Inconsistent Topologies}\label{sec:results:scn-2}

The second scenario of this study combines the results of wing structures with different structural layouts.
Unlike Scenario~1, the fields of the main and auxiliary datasets have inconsistent topologies.
As discussed in Section~\ref{sec:results:scn-1}, the auxiliary analysis is the one using a structural layout with a low-rib count (see Figure~\ref{fig:scn-2:layout}).
It is assumed that this data is readily available from a previous study, and is used to enhance the main data corresponding to a new design study.

\begin{figure}[htb]
    \centering
    \includegraphics{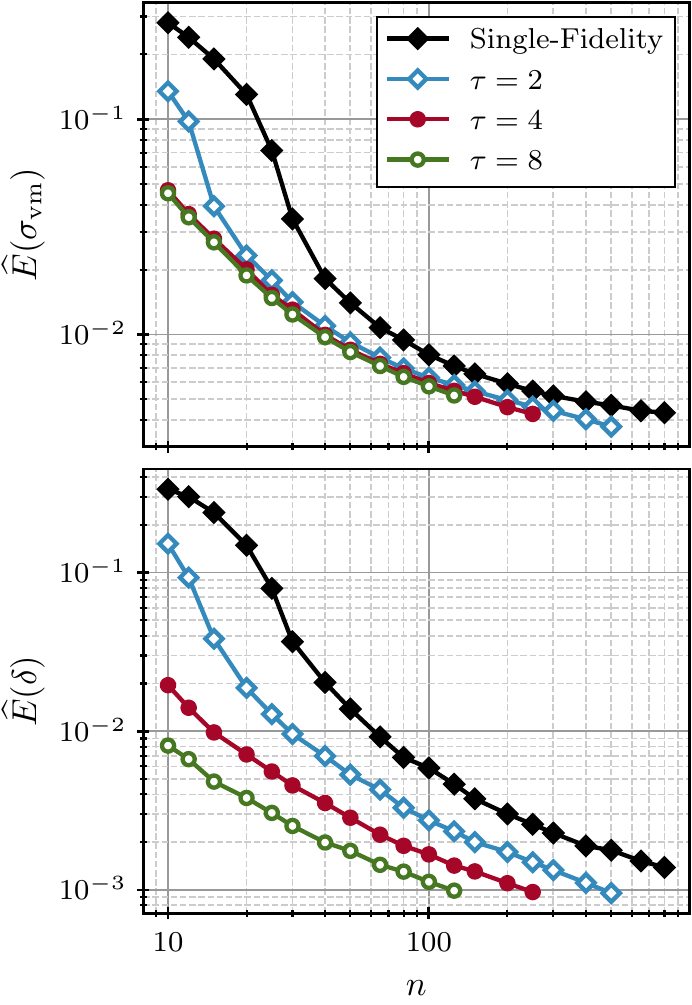}
    \caption{Prediction errors of $\sigma_\text{vm}$ and $\delta$ using the MA-ROM method to combine results from different structural topologies.}
    \label{fig:scn-2}
\end{figure}

The results in Figure~\ref{fig:scn-2} present the prediction error obtained with the MA-ROM method applied to the current scenario.
As in the previous scenario, the normalized error is shown for a range of high-fidelity samples $n$ and multiple multi-fidelity ratios $\tau$.
The error achieved with a single-fidelity approach is also shown for comparison.
The top part of Figure~\ref{fig:scn-2} gives the prediction error for the von-Mises stress field $\sigma_\text{vm}$, while the bottom part shows the error for the structural displacement field.
The numerical values for the cases at $n=100$ are also listed in Table~\ref{tab:scn-2} for an easier comparison.
Unlike in Figure~\ref{fig:scn-1:stress} and~\ref{fig:scn-1:displacement} in Scenario~1, the computational cost is not shown here since the auxiliary data is assumed to already exist.

\begin{table}[]
    \centering
    \caption{Prediction error of the MA-ROM method applied to Scenario~2.}
    \label{tab:scn-2}
    \begin{tabular}{cccc}
    \hline
    $n$ & $m$ & $\widehat{E}(\sigma_\text{vm})$ & $\widehat{E}(\delta)$ \\
    \hline\hline
    100$^*$ & -   & 0.805\% & 0.587\% \\
    100     & 200 & 0.630\% & 0.275\% \\
    100     & 400 & 0.595\% & 0.167\% \\ 
    100     & 800 & 0.574\% & 0.112\% \\
    \hline
    \multicolumn{4}{l}{$^*$\footnotesize{Results for a single-fidelity ROM}}
    \end{tabular}
\end{table}

When compared to the results reported in Section~\ref{sec:results:scn-1}, the current results exhibit trends in agreement with those of Scenario~1.
Once again, the normalized error decreases monotonically with additional training data and adding auxiliary data produces better results than using the main dataset only.
For $n=100$, the prediction error is up to 28.70\% and 80.92\% smaller for $\sigma_\text{vm}$ and $\delta$ respectively, which is comparable to the results presented in Table~\ref{tab:scn-1}.
Furthermore, the results of Figure~\ref{fig:scn-2} indicate once more that improving the prediction of $\sigma_\text{vm}$ with the MA-ROM method is more challenging than it is for $\delta$.
This reinforces the suspicion that the higher complexity and non-linearity of the von-Mises stress field renders the main and auxiliary fields more difficult to fuse.

The main outcome of the results of Scenario~2 is that combining data with disparate topologies does not appear to alter the performance of the MA-ROM method.
Therefore, the accuracy of a field prediction model can readily be increased using MA-ROM if data from a similar, yet different problem is available even if their field size or topology are inconsistent.
For instance, this could be useful if a ROM was previously trained for an earlier iteration of the design, and is no longer adequate due to changes to the structural layout.
Consequently, the MA-ROM method gives renewed value to old, discarded data by using them to enhance the prediction of models trained on new, expensive data.

\subsection{Field Visualization}

\begin{figure}[htb]
    \centering
    \includegraphics{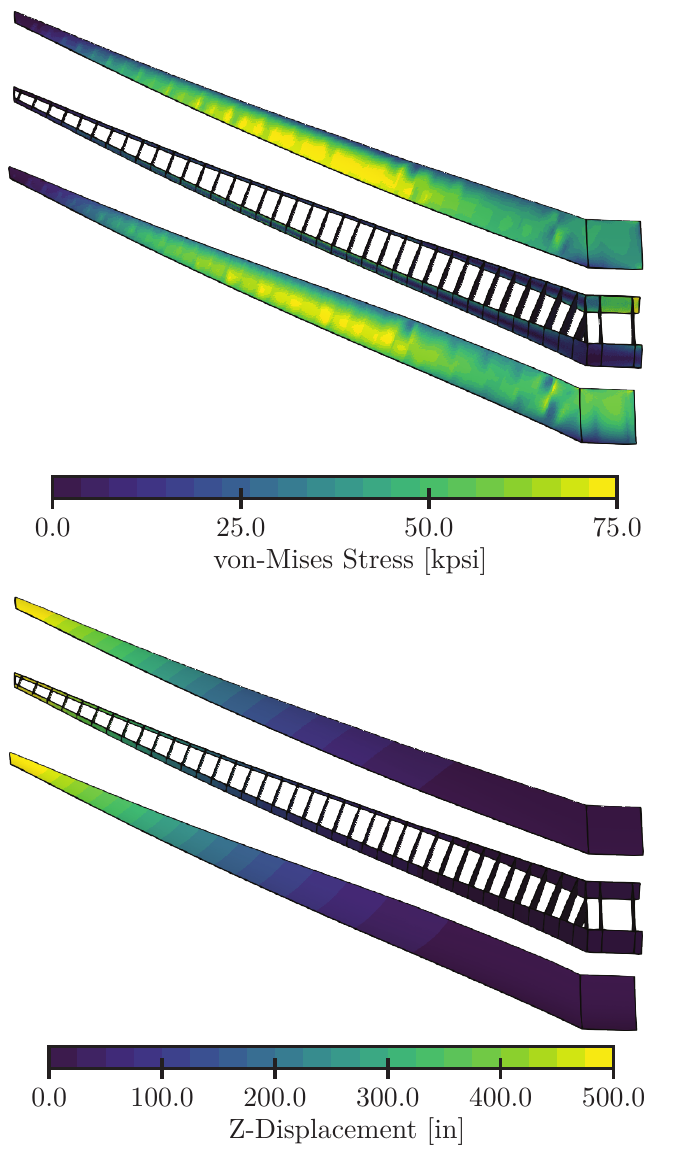}
    \caption{Exact simulation results of the von-Mises stress and the $z$ displacement fields for the test case with the highest prediction error.}
    \label{fig:worst_point}
\end{figure}

\begin{figure*}[p]
    \centering
    \includegraphics{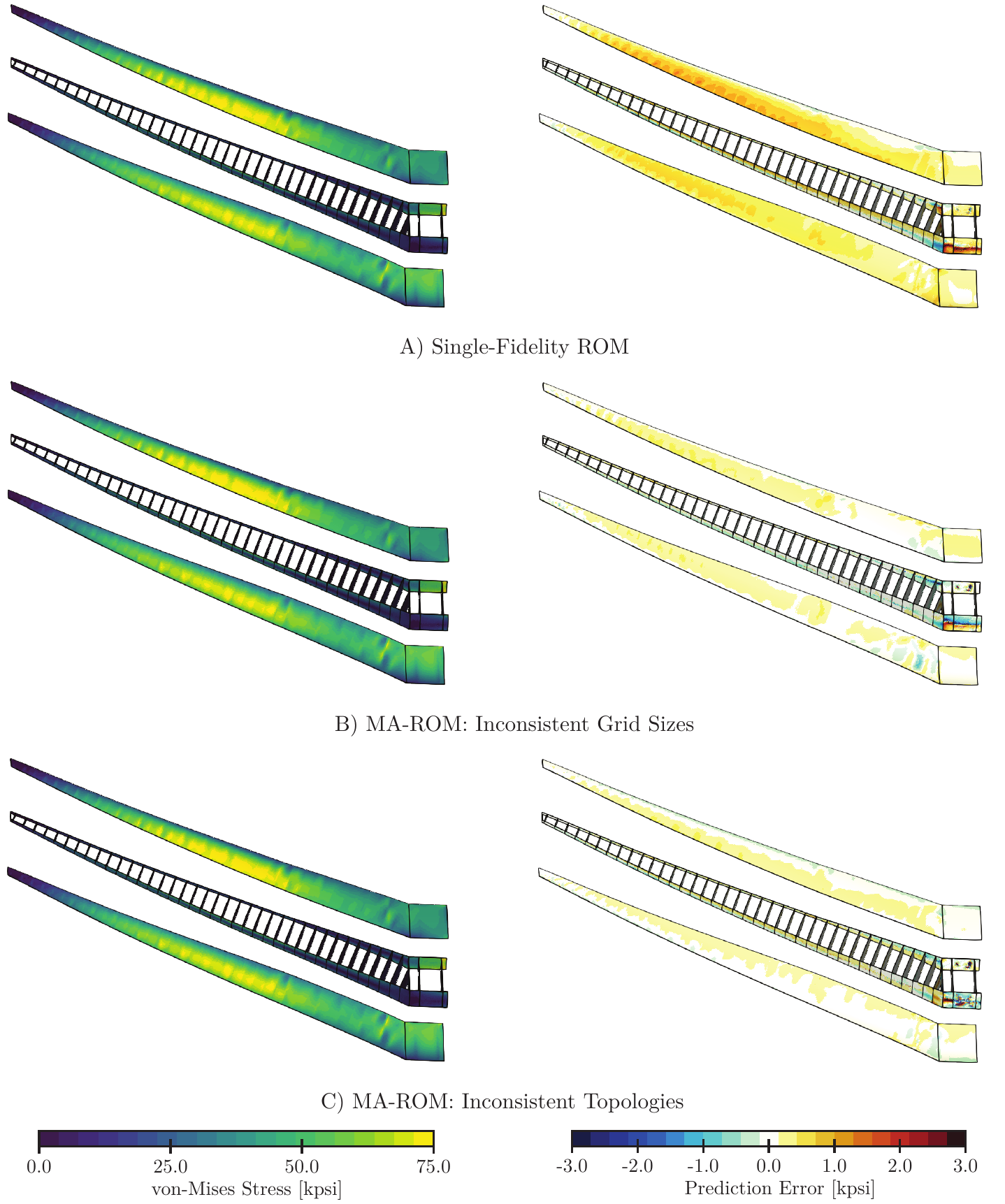}
    \caption{Prediction results of the von-Mises stress field for the test case with the highest prediction error. Shown results consider a single-fidelity ROM model~(A), MA-ROM model combining inconsistent grid sizes~(B), and a MA-ROM model combining inconsistent structural topologies~(C).}
    \label{fig:stress_visuals}
\end{figure*}

\begin{figure*}[p]
    \centering
    \includegraphics{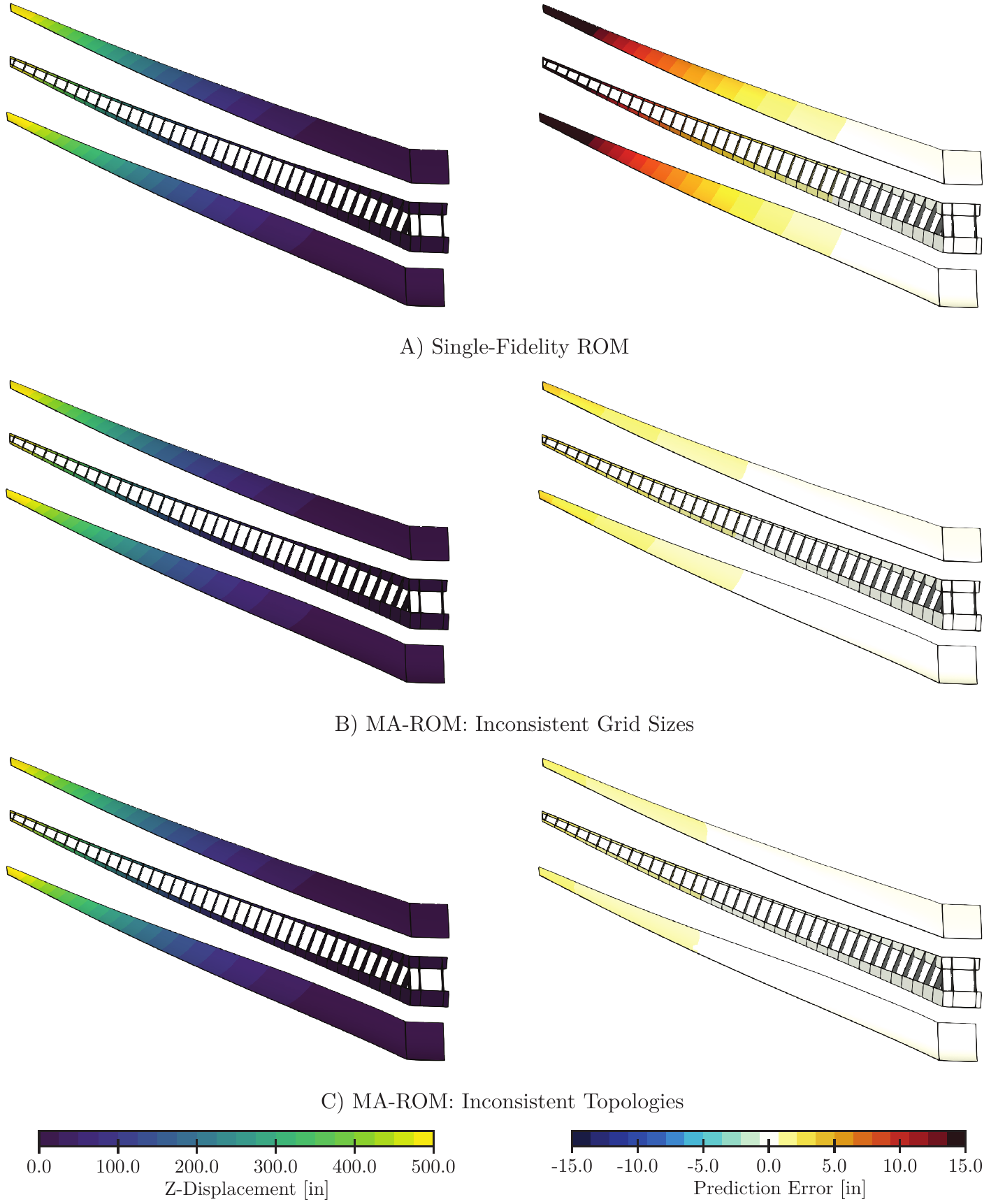}
    \caption{Prediction results of the $z$ displacement field for the test case with the highest prediction error. Shown results consider a single-fidelity ROM model~(A), MA-ROM model combining inconsistent grid sizes~(B), and a MA-ROM model combining inconsistent structural topologies~(C).}
    \label{fig:displacement_visuals}
\end{figure*}

This section presents a few key figures for a visual depiction of the predictive accuracy provided by the MA-ROM.
For reference, Figure~\ref{fig:worst_point} shows the actual solution for the test point with the largest observed prediction error.
Recollect that single- and multi-fidelity ROMs are constructed to predict both the stress and the displacement fields.
It should be noted that the test case presented corresponds to a design that would normally be infeasible due to the high stress and displacement values computed.
However, other than being an extreme design, this does not impact the ROM models as they merely reproduce the analysis results.
Figures~\ref{fig:stress_visuals} and~\ref{fig:displacement_visuals} then present the predicted fields and the corresponding prediction errors for both quantities of interest.
The results for both multi-fidelity scenarios are shown, as well as the results from a single-fidelity approach for comparison sake.

Comparing Figures~\ref{fig:stress_visuals} and~\ref{fig:displacement_visuals} to Figure~\ref{fig:worst_point} shows that all ROM methods can recover the actual field of interest with a reasonable accuracy.
Yet, an inspection of the error fields reveals that for the prediction of both the stress and the displacement fields, the MA-ROM outperforms the single-fidelity POD-based ROM even for the worst prediction in the test set.
These visualizations also show that the error tends to be higher near regions of high stress (e.g., the wing root) or large displacement (e.g., the wing tip)
One should further note that the error levels for both multi-fidelity scenarios have similar magnitudes.
These observations support previous discussions and reinforce MA-ROM's efficacy when applied to high-fidelity structural analyses with grid inconsistencies and differences in topology.


\section{Conclusion}\label{sec:conclusion}

Reduced-order models are a popular solution to create a computationally cheap model useful for many-query applications.
The usefulness and adoption of these \emph{quick-to-evaluate} models have increased further due to multi-fidelity techniques that use sparse high-fidelity and abundant low-fidelity data.
Such models tackle the trade-off between predictive accuracy and the cost of generating a sufficient amount of expensive training data. 

The structural design and sizing of aircraft components at the preliminary design phase is a scenario of many-query applications. A feature of this problem is the necessity to accurately predict the high-dimensional field responses. The field responses from physics-based structural simulation models of different fidelity have disparate representations. These inconsistencies are an obstacle to the use of contemporary approaches for multi-fidelity ROMs. 

In this work, we proposed the use of a recent multi-fidelity ROM method that is capable of merging fields with inconsistent dimensionalities, topologies, and features. The method projects the low- and high-fidelity data to their respective reduced subspaces. Then, a Procrustes analysis is used to align the two manifolds. The alignment allows for the two datasets to be combined into a multi-fidelity ROM. 

The proposed method is demonstrated on two structural scenarios that have multi-fidelity fields of inconsistent representations. The first scenario tackled inconsistent grid sizes by combining data from a fine structural mesh with that from a coarse structural mesh. The second scenario tackled inconsistent topologies by combining data from a high-rib count structure with that from a low-rib count structure. 

Results showed that the method could improve the accuracy by up to 62.6\% when compared to a single-fidelity ROM trained with a similar computational budget.
While lower errors were observed for the prediction of both the von-Mises stress and the structural displacement fields, the MA-ROM was noticeably more suited to the evaluation of the latter.
Indeed, for a set of high-fidelity samples, the proposed method would become more accurate with additional low-fidelity results, but when applied to the stress field, the lowest cost for a specified accuracy is found at $\tau = 4$.
The reduced effectiveness of the MA-ROM method for the prediction of the von-Mises is explained by the higher complexity and non-linearity of that field when compared to the structural displacement.

An extension of the present work is to explore creating a multi-fidelity ROM combining results from beam and shell models.
The computationally efficient beam model would provide low-fidelity data owing to its inaccuracy at capturing the torsional response due to the ribs' absence. 
On the other hand, the computationally inefficient, and generally more accurate, shell model would provide the high-fidelity data.
The authors also intend to extend the current method to adaptive sampling scenarios.
This would enable even greater cost reduction in the context of surrogate-based optimization.



\backmatter

\section*{Declarations}

\bmhead{Supplementary information}

This article has the following accompanying material that will be made available after receiving peer reviews: 1) Computer program that was developed for constructing the predictive model; 2) Data that was used to construct the models.

\bmhead{Funding}

No funding, grants, or other monetary support was received. 

\bmhead{Conflict of Interest/Competing Interests}

The authors declare that there is no conflict of interest.

\bmhead{Availability of Data and Materials}

The data and accompanying material used to generate results presented in this work will be made available as electronic supplementary material. 

\bmhead{Code Availability}

The computer program that implements the method presented in this work will be made available through a repository link. 

\bmhead{Financial/Non-Financial Interests}

The authors have no relevant financial or non-financial interests to disclose. 

\bibliographystyle{sn-aps.bst}
\bibliography{Parts/references.bib}


\end{document}